\pdfoutput=1

\documentclass[11pt]{article}

\usepackage[preprint]{acl}

\usepackage{times}
\usepackage{latexsym}

\usepackage[T1]{fontenc}

\usepackage[utf8]{inputenc}

\usepackage{microtype}

\usepackage{inconsolata}

\usepackage{graphicx}

\usepackage{listings}
\usepackage{booktabs,multirow,makecell}
\usepackage{graphicx}

\newcommand{\myname}[0]{\textsc{DeepSpecs}}
\newcommand{\bbb}[1]{\noindent\textbf{#1}}

\title{DeepSpecs: Expert-Level Questions Answering in 5G}

\author{
 \textbf{Aman Ganapathy Manvattira\thanks{~Equal contribution.}},
 \textbf{Yifei Xu\footnotemark[1]},
 \textbf{Ziyue Dang},
 \textbf{Songwu Lu}
\\
University of California, Los Angeles
\\
amanvatt02@g.ucla.edu,
\{yxu, ziyue.dang, slu\}@cs.ucla.edu
}

\begin{document}
\maketitle
\begin{abstract}
5G technology enables mobile Internet access for billions of users. Answering expert-level questions about 5G specifications requires navigating thousands of pages of cross-referenced standards that evolve across releases. Existing retrieval-augmented generation (RAG) frameworks, including telecom-specific approaches, rely on semantic similarity and cannot reliably resolve cross-references or reason about specification evolution. We present \textsc{DeepSpecs}, a RAG system enhanced by structural and temporal reasoning via three metadata-rich databases: SpecDB (clause-aligned specification text), ChangeDB (line-level version diffs), and TDocDB (standardization meeting documents). \textsc{DeepSpecs} explicitly resolves cross-references by recursively retrieving referenced clauses through metadata lookup, and traces specification evolution by mining changes and linking them to Change Requests that document design rationale. We curate two 5G QA datasets: 573 expert-annotated real-world questions from practitioner forums and educational resources, and 350 evolution-focused questions derived from approved Change Requests. Across multiple LLM backends, \textsc{DeepSpecs} outperforms base models and state-of-the-art telecom RAG systems; ablations confirm that explicit cross-reference resolution and evolution-aware retrieval substantially improve answer quality, underscoring the value of modeling the structural and temporal properties of 5G standards.

\end{abstract}

\section{Introduction}

5G technology enables mobile and wireless Internet access for billions of users globally. At the foundation of this ecosystem lies the 3GPP standard, a continuously evolving set of specifications that define every technical detail of 5G systems. These documents span thousands of pages per release and cover diverse aspects such as architecture, protocols, and procedures. They are vital to the entire telecom workflow, including design, implementation, testing, and operations~\cite{3gpp_overview}.
Practitioners such as system architects, field engineers, and vendor developers frequently need to answer complex technical questions grounded in these specifications. For example, they may ask “What are the conditions required to support Feature X?” or “When and why was Feature Y introduced?” Answering such questions typically involves navigating multiple cross-referred documents, tracing changes across different releases, and understanding the rationale behind design decisions. This process is time-consuming and requires substantial domain expertise~\cite{chen2022seeing, rodriguez2025technical}.

Recent advances in large language models (LLMs) and retrieval-augmented generation (RAG) have emerged as promising tools for automating technical question answering~\cite{lewis2020retrieval, guu2020retrieval, izacard2023atlas}.
Recent efforts have adapted these approaches to the telecom domain. Systems such as Tele-LLMs~\cite{maatouk2024tele}, Telco-RAG~\cite{bornea2024telco}, and Chat3GPP~\cite{huang2025chat3gpp} incorporate domain-specific retrieval or fine-tuning on 3GPP specifications. However, these systems largely follow vanilla RAG pipelines without considering the unique inter-document relationships that span across the 3GPP corpus. Two core challenges remain when applying these methods to 5G specifications.
\textit{(1) Dense cross-references:} 5G specifications are designed for modularity and precision. Instead of repeating content, for readability they refer extensively to other clauses or documents via specification and clause IDs~\cite{qualcomm_3gpp_basics, 3gpp_spec_tech}. As a result, critical definitions or conditions are often located outside of semantically similar text, requiring accurate reference resolution beyond what standard semantic retrieval can provide.
\textit{(2) Implicit reasoning behind the evolution of specifications:} As 5G specifications evolve incrementally over many releases, each version defines only what the intended behavior is at that point in time. Design rationale and historical context are excluded by intent, as specifications aim to specify "what is" rather than "why it changed." This makes it challenging to trace when and how a feature was introduced, modified, or deprecated without consulting external sources such as version changes and meeting discussions~\cite{baron2018unpacking, chen2022seeing}.

To address these gaps, we present \textbf{\myname{}}, a QA system tailored to the 5G domain. \myname{} extends the RAG paradigm with structural and temporal understanding of 3GPP specifications. At its core, \myname{} integrates three domain-specific, metadata-rich databases: \textit{SpecDB}, which stores aligned and structured specification contents; \textit{ChangeDB}, which tracks line-level changes and associated metadata across releases; and \textit{TDocDB}, which organizes 3GPP meeting documents (TDocs) to support reasoning about design-phase intents. Building on this foundation, \myname{} provides two clause-level capabilities that mimic how a human expert navigates the standards: (i) \textit{cross-reference resolution} via rule-based reference extraction and hybrid retrieval, enabling fine-grained navigation within and across documents; and (ii) \textit{specification-evolution reasoning} by mining clause changes and linking them to corresponding discussions through 3GPP-specific metadata, allowing the system to recall when a feature was introduced, how it evolved, and why.

To evaluate the effectiveness of our method, we curate a real-world dataset of 573 question–answer pairs spanning over 3 years from public telecom forums and blog posts, annotated by three telecom experts. The dataset covers practitioner-facing questions across RAN, core, protocol features, and evolution-related queries, providing a realistic testbed for telecom QA. \myname{} demonstrates consistent gains over strong LLM baselines and a state-of-the-art telecom-specific RAG system. We further design targeted microbenchmarks for cross-reference resolution and specification-evolution reasoning, and validate the effectiveness of \myname{} on both.

The source code for \myname{} is currently available upon email request.

\paragraph{Contributions.}
\begin{itemize}
\item We introduce \myname{}, a novel framework for 5G QA that enables clause-level cross-reference resolution and specification-evolution reasoning, supporting expert-level question answering over technical standards.
\item We present, to our knowledge, the first expert-annotated 5G QA dataset, comprising 573 question–answer pairs collected from real-world practitioner forums and blog posts.
\item Through automated and human evaluation, we show that \myname{} outperforms strong LLM baselines combined with state-of-the-art retrieval, and demonstrate the benefit of cross-reference resolution and specification-evolution reasoning.
\end{itemize}

\section{Related Work}
\textbf{RAG Approach for Advanced QA}: 
Standard RAG systems like Atlas \cite{izacard2023atlas} and IRCoT \cite{trivedi2023interleaving} improve factual consistency through retrieval. However, they struggle with technical specifications like 5G, which require precise resolution of intra- and inter-document references. Standard RAG retrieves isolated text chunks without resolving cross-references or tracking specification evolution across versions. They also lack support for surfacing the rationale behind changes, limiting their utility in domains where accurate interpretation of evolving, interconnected documents is critical. Recent RAG research addresses some gaps: GraphRAG \cite{edge2024local} models text as knowledge graphs; agentic frameworks like Legal Document RAG \cite{zhang2025towards} and FinSage \cite{wang2025finsage} employ multi-agent traversal; Temporal RAG systems such as $E^2RAG$ \cite{zhang2025respecting} and ConQRet \cite{dhole-etal-2025-conqret} support fine-grained reasoning over evolving corpora. However, these have yet to address the technical and versioned complexity of telecom standards. Our system extends this frontier by integrating structural, temporal, and argumentative retrieval to enable evolution-aware, cross-referenced question answering for 5G specifications.

\noindent\textbf{Domain-Specific Telecom QA Systems}: 
Recent advances in LLMs have enabled their use in telecommunications to improve access to the complex LTE/5G standard specifications. Tele-LLMs \cite{maatouk2024tele} are domain-specialized models trained on 3GPP specifications, achieving improvements over general LLMs. 
RAG-based systems such as Chat3GPP \cite{huang2025chat3gpp}, Telco-RAG \cite{bornea2024telco}, and Telco-oRAG \cite{bornea2025telco} employ hybrid dense-sparse retrieval, neural routing and query reformulation for 3GPP document collections. While these systems improve upon basic semantic search, they remain fundamentally vanilla RAG approaches that retrieve text chunks based on semantic similarity without explicit support for cross-reference resolution or specification evolution tracking.
Complementary to these systems, TSpec-LLM \cite{nikbakht2024tspec} provides a comprehensive 3GPP dataset and shows improved QA performance using naive RAG, but its evaluation is limited to multiple-choice questions and lacks support for reference resolution or specification evolution.
In contrast, our system extends this line of work by incorporating structural, temporal, and argumentative retrieval tailored to 3GPP specifications. It explicitly resolves inter-document references and tracks specification evolution over time, which are capabilities essential for answering expert-level telecom questions. We also develop new datasets targeting reference resolution and specification evolution, enabling evaluation beyond multiple-choice formats.

\section{Method}

\subsection{Goal}

Our goal is to build a system that answers expert-level questions over 5G specifications by producing accurate and helpful responses. This requires structural and temporal understanding beyond surface semantics, in particular: (i) \textit{cross-reference resolution}, which follows spec IDs and clause numbers to integrate information scattered across modular, non-redundant documents; and (ii) \textit{specification-evolution reasoning}, which traces when a feature was introduced, how it changed across releases, and why. In line with practitioner practice, our goal emphasizes not only factual accuracy but also explanation helpfulness, reflecting how experts contextualize and justify their answers.

\subsection{Overview}

\myname{} answers expert-level questions over 5G specifications by extending a RAG framework with structural and temporal reasoning. Following human expert practice, it supports clause-level cross-reference resolution and specification-evolution reasoning.

\myname{} operates in four stages: 
\textit{(1) Database construction}, where specification content and TDocs are indexed into three databases: \textit{SpecDB}, serving as the primary context provider enriched with metadata for accurate reference retrieval; \textit{ChangeDB} and \textit{TDocDB}, which capture the temporal evolution of specifications and support reasoning over changes and design decisions; 
\textit{(2) Cross-reference resolution}, which extracts specification and clause IDs using rule-based patterns and retrieves the corresponding clauses from filtered chunks with metadata; 
\textit{(3) Specification-evolution reasoning}, which maps the queried feature to relevant entries in ChangeDB and uses the associated metadata to retrieve design-phase reasoning from TDocDB; and 
\textit{(4) Answer generation}, which produces the final response. 
Each component is described in detail below.

\begin{figure*}[t]
\centering
\includegraphics[width=1\linewidth]{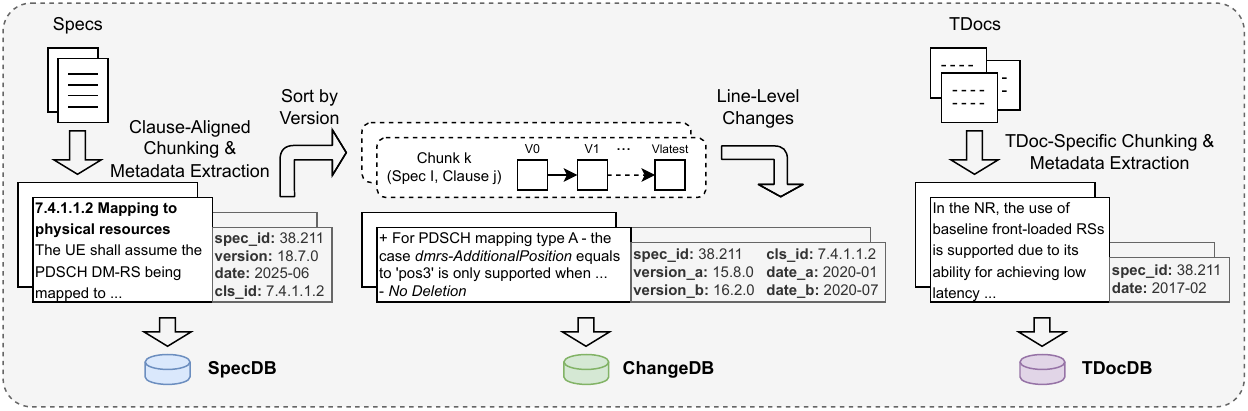}
\caption{Database construction of \myname{}. The system builds three vector databases from 5G specs and TDocs, each embedded with rich metadata: SpecDB provides the backbone context; ChangeDB tracks the temporal evolution of specs; TDocDB stores the reasoning behind each change.}
\label{fig:db}
\end{figure*}

\subsection{Database Construction}

As illustrated in Figure~\ref{fig:db}, we construct three structured databases that serve as the core retrieval sources for \myname{}.

\bbb{SpecDB:}
This database stores the core content of 3GPP specification documents. We download official DOCX-formatted specifications from the 3GPP official archive\footnote{\url{https://www.3gpp.org/ftp/}}. We chunk documents by atomic clauses (e.g., ``7.4.1.1.2 Mapping to physical resource'') and annotated with metadata including specification ID, clause ID, version number, and timestamp, which are extracted based on structural patterns followed consistently across 3GPP specs \cite{openair_walkthrough}. This clause-aligned indexing enables fine-grained retrieval and also serves as the basis for constructing ChangeDB. We employ an LLM for metadata extraction.

\bbb{ChangeDB:}
This database records line-level differences between adjacent versions of each specification clause. Using the clause-aligned chunks derived from SpecDB, we sort all versions of a given clause by timestamp and skip missing versions for robustness consideration. We then compute diffs between each pair of adjacent versions and extract added, removed, or modified lines. Each change entry is annotated with metadata from both versions, including timestamps, and is indexed for temporal reasoning. We also treat the first observed version of a clause as its initial addition, allowing us to track the introduction of new features.

\bbb{TDocDB:}
This database stores 3GPP meeting documents (TDocs), which capture design-phase discussions explaining feature additions, protocol trade-offs, and specification ambiguities. Although such content is essential for understanding the rationale behind changes, it is omitted from the formal specifications for reasons of brevity and standardization \cite{sharetechnote2025, baron2018unpacking}. In this work, we focus on \textit{Change Requests} (CRs), which document the reasoning behind each proposed modification. We collect DOCX-formatted CRs from the official archive. These CRs follow a consistent structure. To process them, we employ an LLM extractor for CR-specific chunking and metadata extraction.
In our parsing strategy for change requests (CRs), we extract content from three key sections: the summary of changes, the reasons for the change, and the consequences of non-approval. Each individual change listed in the summary is segmented into a separate chunk, within which it is then paired with its corresponding rationale and the potential consequences if the change is not adopted.

All three databases are indexed for both exact metadata filtering and dense semantic retrieval using OpenAI's \texttt{text-embedding-3-large} embeddings~\cite{textembedding3}. We use these indices for hybrid retrieval during following reference resolution and evolution reasoning stages.

\begin{figure*}[t]
\centering
\includegraphics[width=1\linewidth]{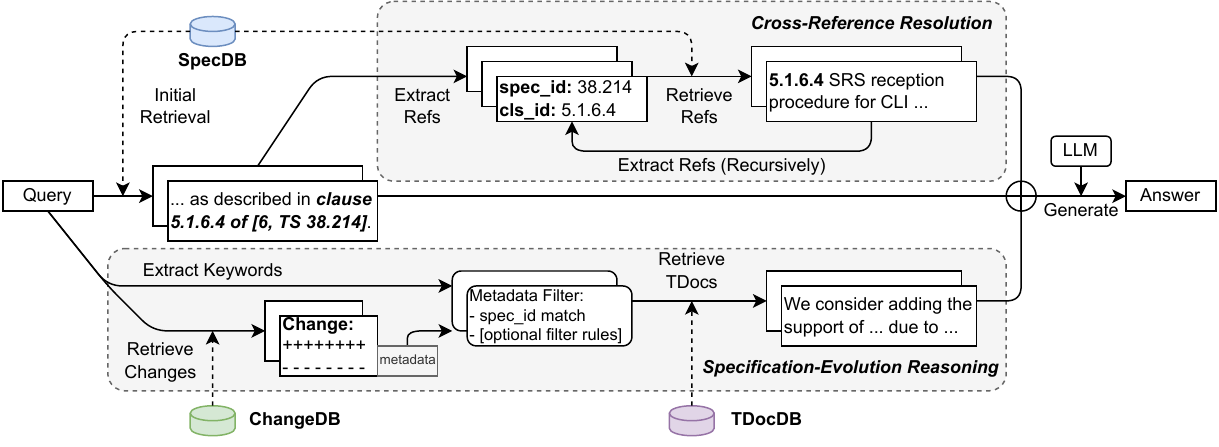}
\caption{The retrieval process of \myname{}. The system leverages the metadata associated with the chunks in each DB to resolve cross-references and trace changes of specs. This allows \myname{} to provide informative context with structural and temporal understanding of 5G specs for question answering.}
\label{fig:retrieval}
\end{figure*}

\subsection{Cross-Reference Resolution}

5G specifications are intentionally modular. To reduce redundancy and maintain formality, clauses frequently cite other clauses within the same document or across different documents. For instance, a clause in TS 38.211 may define a feature but defer key operational conditions to TS 38.214. As a result, answering a single technical query often requires chaining together multiple cross-referred clauses \cite{openair_walkthrough}.

To support this, \myname{} implements clause-level cross-reference resolution, as shown in Figure~\ref{fig:retrieval}. The resolution process consists of the following steps:

\bbb{(1) Initial Retrieval:} Given a user query, we first retrieve top-$k_1$ ranked passages from SpecDB according to semantic similarity. Specifically, we employ HyDE~\cite{gao2023precise} in this phase since practical user queries in the 5G domain often lack sufficient context for effective zero-shot retrieval.

\bbb{(2) Reference Extraction:} From the initially retrieved passages, we extract all cited specification and clause ID pairs \texttt{<spec\_id, cls\_id>}. This is done using regular expressions based on citation patterns commonly found in 3GPP documents, such as “See clause 5.1.6.4 of TS 38.214.” 

\bbb{(3) Reference Retrieval:} Each pair of extracted \texttt{<spec\_id, cls\_id>} is used to locate the corresponding clause in SpecDB by matching its spec ID and clause ID through metadata lookup. If the retrieved chunk contains additional references, the system recursively resolves them to build a deeper context chain. To avoid over-expansion, we use a configurable maximum recursion depth. This process enables the system to construct a retrieval trace that reflects the true dependency structure of the initially retrieved passage.

\bbb{(4) Reference Ranking:} All referenced chunks, including those obtained recursively, are re-ranked according to their semantic relevance to the original question. The top-$k_2$ chunks are returned for final answer generation.

\subsection{Specification-Evolution Reasoning}

Many expert-level queries require understanding how a feature has changed across specification versions and why certain changes were introduced. While 3GPP specifications describe the current intended behavior with precision, they omit the rationale behind changes for conciseness and neutrality \cite{baron2018unpacking}. This rationale is often recorded instead in 3GPP meeting documents (TDocs), which capture discussions, debates, and design motivations during the standardization process.

As illustrated in Figure~\ref{fig:retrieval}, \myname{} leverages the TDocs and traces the evolution of a feature with the following steps:

\bbb{(1) Direct Extraction:}
\myname{} first attempts to extract explicit mentions of specification IDs from the user query (practically, users often name a spec directly). The extracted spec IDs feed a later metadata-based filter. We employ an LLM extractor for this step.

\bbb{(2) Change Retrieval:}
If no specification is mentioned explicitly, \myname{} queries \textsc{ChangeDB} using dense semantic retrieval (along with HyDE) to locate the most relevant change entry. This step discovers additions, removals, or modifications related to the queried feature.

\bbb{(3) Metadata-Based Filtering:}
Using the metadata (e.g., date, spec ID) of the identified change, \myname{} maps the query or retrieved change (if any) to a narrowed set of candidate TDocs. While our current implementation matches by spec ID only, which is sufficiently accurate for CRs, the \myname{} framework also supports filtering with additional metadata (e.g., working group, feature tags) and rules when other TDoc types are included.

\bbb{(4) TDoc Ranking:}
All filtered TDocs from \textsc{TDocDB} are re-ranked according to their semantic relevance to the HyDE aligned hypothetical document generated from the original query. The top-$k_3$ chunks are returned for final answer generation.

This hybrid strategy, supported by ChangeDB, not only grounds the query-to-TDoc mapping in explicit specification metadata but also enhances the interpretability and transparency of retrieval compared to purely semantic methods, which is especially valuable for practitioners.

\subsection{Answer Generation}

After retrieving relevant specification clauses, resolving cross-references, and identifying supporting TDocs, \myname{} assembles a generation prompt for a general-purpose language model to synthesize the final answer. From the stage of initial retrieval, reference resolution, and specification-evolution reasoning, it selects the top-$k_1$, $k_2$, and $k_3$ chunks, respectively, ranked by semantic similarity.

\section{Datasets}
There are very few established benchmarks on telecom QA and most emphasize surface-level factual queries, rely on synthetic QA collections, and often restrict evaluation to multiple-choice problems~\cite{maatouk2024tele,huang2025chat3gpp,bornea2024telco,bornea2025telco,nikbakht2024tspec}. We address these gaps by constructing two new datasets that capture (i) real-world practitioner information needs and (ii) reasoning behind the evolutions of specifications.

\subsection{Real-World QA Dataset}
To evaluate system performance in realistic professional settings, we curate a dataset of question–answer pairs that reflect genuine practitioner information needs. The data are drawn from publicly available forums and educational resources where telecommunications professionals actively exchange knowledge on 5G, ensuring alignment with engineering and instructional use cases.

\paragraph{Data Collection.}
We collect QA data from two widely used sources spanning a three-year period from August 2022 to August 2025. \textit{telecomHall}~\cite{telecomHall} is a long-standing global telecommunications community (established in 1999) where practitioners discuss technical challenges. We retain only explicit question–answer exchanges that are technically accurate, complete, and well-formed. \textit{ShareTechnote}~\cite{ShareTechnote} is an educational resource with extensive technical documentation on cellular communications, with a strong emphasis on 5G. From 236 webpages, we formulate precise questions and extract answers directly from the material.
The questions are categorized into five technical domains (detailed taxonomy in Appendix~\ref{appendix:real_world_qa}).

\paragraph{Human Annotation.}
Three volunteers with advanced backgrounds in mobile networks and cellular systems review each QA pair to validate technical correctness, clarity, and relevance. The curation, data collection, and verification process require approximately 120 hours of annotation effort across a 3-month collection period.

\subsection{CR-Focused QA Dataset}

To support targeted evaluation of \myname{} specification-evolution reasoning, we construct another QA dataset of derived from real 3GPP specification changes with rich supervision signals.

\paragraph{Data Collection.}
We collect approved Change Requests from 3GPP specifications spanning Releases 17 and 18. The collection yield 997 CR documents, with the majority addressing physical layer procedures, channels, and multiplexing, along with dual connectivity enhancements. 
Each approved CR is retrieved with its associated TSG documentation package containing the complete change proposal, rationale, technical details, and impact analysis.

\paragraph{QA Generation.}
We use a combination of LLM and human verification to generate the QA pairs. We first extract three key fields from each CR: summary of change, reason for change, and consequences if not approved. To ensure quality, we filter out trivial CRs where all three fields contain only brief descriptions (fewer than 200 words each), typically representing minor editorial corrections lacking sufficient technical depth.
For substantive CRs, we employ an LLM to generate question-answer pairs following a structured prompt (detailed in Appendix~\ref{appendix:cr_qa_prompt}). 
All generated QA pairs are manually reviewed to verify technical accuracy, question practicality, and answer quality.

\subsection{Dataset Characteristics.}
Table~\ref{tab:qa_stats} summarizes statistics of the final collected real-world and CR-focused QA datasets.
\begin{table}[t]
\centering
\resizebox{\columnwidth}{!}{%
\begin{tabular}{lccc}
\toprule
\textbf{Source} & \textbf{\#} & \textbf{Avg. Q Len} & \textbf{Avg. A Len} \\
\midrule
\multicolumn{4}{l}{\textit{Real-world QA Dataset}} \\
telecomHall     & 102 & 46.9 tokens & 126.9 tokens \\
ShareTechnote   & 471 & 30.3 tokens & 102.7 tokens \\
\textbf{Total}  & 573 & 33.2 tokens & 107.0 tokens \\
\midrule
\multicolumn{4}{l}{\textit{CR-focused QA Dataset}} \\
3GPP Change Requests & 350 & 45.4 tokens & 287.7 tokens \\
\bottomrule
\end{tabular}%
}
\caption{Statistics of the QA datasets, including real-world sources and CR-based pairs.}
\label{tab:qa_stats}
\end{table}

\section{Experiments}

We evaluate \myname{} on its ability to answer expert-level questions about 5G specifications. Specifically, we ask:

\begin{enumerate}
    \item How well does \myname{} answer practical questions encountered by practitioners, compared to the state-of-the-art baselines?
    \item How does semantic retrieval differ from cross-reference resolution performed by experts?
    \item How well does \myname{} perform specification-evolution reasoning on tasks targeting evolution-related reasoning?
\end{enumerate}

\subsection{Experimental Setup}

\subsubsection{Retrieval Databases}
For \textsc{SpecDB}, we collect all Release~17 and 18 3GPP technical specifications, following prior work~\cite{huang2025chat3gpp}. In total, we obtain 2137 specifications as input for database construction.  
We download Change Requests from 3GPP Releases 17 and 18 that satisfy "approved" status requirements, yielding 997 TDocs in total.  

\subsubsection{Baselines}
We compare \myname{} against two categories of baselines:
\bbb{(1) Base Model:}
Direct prompting of the base LLMs without any retrieval context. For consistency, we use the same answer-generation prompt, 
but replace the context section with ``\texttt{NO CONTEXT}.''
\bbb{(2) Chat3GPP:}
A state-of-the-art RAG system tailored for telecom-domain question answering~\cite{huang2025chat3gpp} employs hybrid retrieval, specialized chunking, and efficient indexing, and demonstrates superior performance over existing methods. Nevertheless, Chat3GPP still largely follows the vanilla RAG paradigm. We use it as our primary baseline to demonstrate the limitations of standard RAG approaches on expert-level 5G questions.

For both baselines, we test a collectin of generation backends, including GPT-4o/4.1/4.1 mini~\cite{openai2024gpt4o, openai2025gpt41}, Qwen3-4/8/14/32B~\cite{yang2025qwen3}, and Claude 3.5 Haiku~\cite{anthropic2024claude3_5haiku}, all using their default or recommended hyperparameter settings.

\subsubsection{Evaluation Metrics}

\paragraph{Pairwise Win Rate.}
We employ an LLM evaluator validated with human evaluation to compute head-to-head win rates between system outputs. The evaluator is given the question, a gold answer, and two candidate answers, and is instructed to select the preferred one based on accuracy and helpfulness. The evaluation prompt is provided in Appendix~\ref{appendix:pairwise}. To mitigate positional bias, evaluation is repeated with reversed answer order.  
We validate the alignment with human evaluators by asking domain experts to re-judge a subset of 144 pairs. The LLM-based judgments are aligned with human assessments in 92.4\% of cases, with most disagreements occurring in close comparisons.

\paragraph{Rubric-Based Scoring.}
To complement pairwise win rates with absolute quality scores, we design question-specific rubrics. For each QA pair, rubrics are first generated with an LLM, then curated and validated by domain experts. During evaluation, an LLM applies these rubrics to score each candidate answer along multiple dimensions. The rubric design and evaluation prompts are provided in Appendix~\ref{appendix:rubrics}.

We provide more implementation details for our experiments in Appendix~\ref{app:implementation}.

\subsection{Results on Real-World QA}

Table \ref{tab:realworld_results} summarizes results across multiple model families. For \myname{}, we fix retrieval parameters ($k_1{=}4$, $k_2{=}3$, $k_3{=}3$), and have Chat3GPP return the top-10 chunks to ensure that both \myname{} and Chat3GPP return the same amount of context.  We report both win rates and output lengths to ensure that observed improvements are not confounded by verbosity bias.

\paragraph{Retrieval consistently helps, with \myname{} providing the strongest gains.}
Both Chat3GPP and \myname{} improve over base models, underscoring the importance of retrieval augmentation for navigating 3GPP specifications. While Chat3GPP offers decent improvements, it remains limited by its inability to resolve cross-references or track evolutions of specifications. In contrast, \myname{} leverages structural- and evolution-aware retrieval to deliver consistent gains across generator models and question categories, showing robustness regardless of model size or output length.
Rubric-based evaluation provides further confirmation of these findings. With GPT-4.1, \myname{} achieves a mean score improvement of 0.62 points on a 5-point scale (p<0.0001, Wilcoxon signed-rank test, dz=0.41), while against GPT-4.1-mini, the improvement is 0.60 points (p<0.0001, dz=0.40). Detailed results on per-category win rates and rubric-based scoring are provided in Appendix~\ref{appendix:real_world_qa}.

\paragraph{Gains are larger on certain weaker models, but remain meaningful for stronger ones.}
The results indicate that even high-capacity models benefit from explicit grounding in structural and temporal aspects of the specifications. 
Given the rapid evolution of 3GPP standards, such grounding becomes especially valuable for maintaining accurate and helpful QA along the time.

\begin{table}[t]
\centering
\caption{Comparative performance on the real-world QA dataset. Both Chat3GPP and \myname{} return the top-10 chunks. For Base GPT-4o, the win rate is fixed at 50\%. }

\resizebox{\columnwidth}{!}{
\begin{tabular}{llcc}
\toprule
\textbf{Model} & \textbf{Method} &
\makecell{\textbf{Length}\\\textbf{(tokens)}} &
\makecell{\textbf{Win Rate}\\\textbf{vs. GPT-4o (\%)}}
\\
\midrule
\multirow{3}{*}{GPT-4o}
  & Base Model   & 295 & (50) \\
  & Chat3GPP     & 295 & 62.5 \\
  & \textbf{\myname{}}    & 301 & \textbf{68.1} \\
\midrule
\multirow{3}{*}{Qwen3-4B}
  & Base Model   & 307 & 34.4 \\
  & Chat3GPP     & 427 & 67.3 \\
  & \textbf{\myname{}}    & 415 & \textbf{71.2} \\
\midrule
\multirow{3}{*}{Qwen3-8B}
  & Base Model   & 346 & 50.8 \\
  & Chat3GPP     & 490 & 74.4 \\
  & \textbf{\myname{}}    & 483 & \textbf{79.8} \\
\midrule
\multirow{3}{*}{Qwen3-14B}
  & Base Model   & 331 & 64.7 \\
  & Chat3GPP     & 459 & 78.9 \\
  & \textbf{\myname{}}    & 460 & \textbf{83.1} \\
\midrule
\multirow{3}{*}{Qwen3-32B}
  & Base Model   & 373 & 73.6 \\
  & Chat3GPP     & 499 & 82.9 \\
  & \textbf{\myname{}}    & 497 & \textbf{88.0} \\
\midrule
\multirow{3}{*}{Claude 3.5 Haiku}
  & Base Model   & 268 & 48.6 \\
  & Chat3GPP     & 278 & 56.6 \\
  & \textbf{\myname{}}    & 278 & \textbf{59.4} \\
\midrule
\multirow{3}{*}{GPT-4.1 mini}
  & Base Model   & 395 & 82.0 \\
  & Chat3GPP     & 424 & 87.8 \\
  & \textbf{\myname{}}    & 442 & \textbf{90.1} \\
\midrule
\multirow{3}{*}{GPT-4.1}
  & Base Model   & 427 & 87.2 \\
  & Chat3GPP     & 494 & 93.7 \\
  & \textbf{\myname{}}    & 511 & \textbf{95.4} \\
\bottomrule
\end{tabular}
}
\label{tab:realworld_results}
\end{table}

\subsection{Evaluating Cross-Reference Resolution}
We quantify the gap between semantic retrieval and cross-reference resolution using a microbenchmark over 10 specifications (TS~38.181–TS~38.304 from release 18). In total, we sample 547 chunks and extract 903 cross-references. These are filtered with an LLM helpfulness judge (Appendix~\ref{appendix:helpfulness_filtering}) and validated by experts to form a set of helpful references, on which we calculate the precision, recall, and f1 score. We compare the retrievals of (a) Chat3GPP and (b) \myname{}’s standalone cross-reference resolver to this set of helpful references. To ensure a fair comparison, we fix retrieval parameters ($k_1{=}3$, $k_2{=}2$, $k_3{=}0$) for \myname{} and have Chat3GPP return the top-5 chunks. Table~\ref{tab:crossref_results} shows substantially higher precision, recall and F1 for \myname{}, indicating the limitations of semantic-only retrieval on interlinked specs and the effectiveness of \myname{}'s cross-reference resolution method.

\begin{table}[t]
\centering
\small
\caption{Cross-reference resolution microbenchmark performance. Precision, recall, and f1 are calculated on the filtered set of "helpful" references. }
\begin{tabular}{lccc}
\toprule
\textbf{Method} & \textbf{Precision} & \textbf{Recall} & \textbf{F1} \\
\midrule
Chat3GPP & 0.0709 & 0.2334 & 0.1031 \\
\textbf{\myname{}} & \textbf{0.1876} & \textbf{0.7055} & \textbf{0.2837} \\
\bottomrule
\end{tabular}
\label{tab:crossref_results}
\end{table}

\subsection{Evaluating Spec-Evolution Reasoning}
We isolate the contribution of evolution reasoning using a CR-focused QA dataset. To control for confounds, we disable reference resolution and fix retrieval parameters ($k_1{=}5$, $k_2{=}0$, $k_3{=}5$), ensuring \myname{} and Chat3GPP retrieve the same number of chunks. Table~\ref{tab:evo_results} reports GPT-4o results. Relative to both the base model and Chat3GPP, \myname{} (with reference resolution disabled) still achieves significantly higher win rates while producing outputs of comparable length. Results for additional models are provided in Appendix~\ref{appendix:evol_full} and show the same trend. Rubric-based evaluation reinforces these findings. On this CR-focused subset, \myname{} achieves a mean quality score of 3.05, compared to 2.48 for Chat3GPP and 2.26 for the base model (all differences significant at p<0.0001, detailed scores in Appendix~\ref{appendix:evol_full}).

\begin{table}[t]
\centering
\caption{Comparative Performance on the CR-Focused QA Dataset. Even with reference resolution disabled, \myname{} outperforms the base model and Chat3GPP at comparable output lengths.}
\resizebox{\columnwidth}{!}{
\begin{tabular}{llcc}
\toprule
\textbf{Model} & \textbf{Method} &
\makecell{\textbf{Length}\\\textbf{(tokens)}} &
\makecell{\textbf{Win Rate}\\\textbf{vs. GPT-4o (\%)}}
\\
\midrule
\multirow{3}{*}{GPT-4o}
  & Base Model   & 273 & (50) \\
  & Chat3GPP     & 265 & 59.3 \\
  & \textbf{\myname{}}    & 269 & \textbf{77.0} \\
\bottomrule
\end{tabular}
}
\label{tab:evo_results}
\end{table}

\section{Conclusion}

We introduce \myname{}, a QA framework infused with the structural and temporal understanding of 3GPP documents. \myname{} performs clause-level cross-reference resolution and specification-evolution reasoning, which are capabilities missing in standard RAG but essential for accurate and helpful question answering in 5G. Evaluations on real-world questions and targeted microbenchmarks demonstrate consistent gains over strong LLM and RAG baselines, with both structural and temporal components contributing to performance.

\section*{Limitations}

\paragraph{Coverage of Specifications and TDocs.}
Our study focuses on English NR (3GPP) specifications, primarily Releases~17--18, and question types skewed toward RAN/PHY procedures. For TDocs, we currently include only CR-type documents. We have not evaluated on other TDoc types, earlier or future releases, or multilingual corpora, so our claims should not be over-generalized.

\paragraph{Structural Assumptions on Documents.}
The method assumes consistent clause numbering, machine-parseable cross-references, and semi-templated Change Requests (CRs). While this holds for most 3GPP documents, exceptions do occur (e.g., atypical numbering, unusual references, or editorial inconsistencies). Such cases can reduce the reliability of linking and reasoning across documents.

\paragraph{Dependence on LLM-Based Processing.}
Our pipeline makes extensive use of large language models for chunking, extraction, and metadata handling. This introduces potential errors and computational or API calling cost. However, most of the processing is one-time or incremental with new releases, and we found that budget-oriented models are sufficient for these tasks. Still, downstream accuracy ultimately depends on model quality, which may vary with version updates.

\paragraph{Evaluation.}
(1) Our annotated datasets are limited in size due to the need for careful expert curation and limited number of high-quality data sources. This constraint may affect statistical significance, and performance estimates should be interpreted with caution.
(2) For our evaluation, some evaluation metrics rely on LLM-based judges or scorers. Although we include human validation, LLM evaluators can reward fluency or plausible reasoning over our desired factors, and prompt bias may persist. We view LLM-based evaluation as a useful complement rather than a full substitute for human evaluation.

\paragraph{Potential Risks.}
Our system generates fluent, citation-backed outputs, but answers may still be incomplete or misleading if context is missed. Over-reliance on such outputs without cross-checking against the specifications could encourage misinterpretation or non-compliant implementations.
As our work is based on public 3GPP standards, direct risks related to privacy, security, or fairness are minimal. We encourage deployment with safeguards such as citation enforcement, uncertainty reporting, and access controls.

\bibliography{custom}
\appendix
\section{Prompt Templates}

\subsection{Pairwise Win Rate}
\label{appendix:pairwise}
We use the following prompt for the pairwise evaluator:
\begin{lstlisting}
# Task
You are an expert in the 5G domain. You will be provided with a 5G-related question, an expert-written reference answer, and two candidate answers to evaluate. Your task is to determine which candidate answer is better based on the reference answer and the following criteria:
- **Answer Accuracy**: How factually correct is each candidate answer compared to the reference answer? Are there any inaccuracies, omissions, or misleading statements?
- **Explanation Helpfulness**: How well does each candidate answer explain the reasoning behind itself? Does it provide useful context, background information, or insights that enhance understanding of the answer?

# Question
{question}

# Reference Answer
{ground_truth}

# Candidate Answers
- Answer A: {answer_a}
- Answer B: {answer_b}

# Instructions
Provide brief reasoning based on the criteria above, followed by your final decision in the format below:
```
Reasoning: <Your brief reasoning>
Judgment: <"Answer A" or "Answer B">
```
\end{lstlisting}

\subsection{Rubric-based scoring}
\label{appendix:rubrics}
We use the following prompt to generate the initial question-specific for each QA pairs:
\begin{lstlisting}
You are an expert in 5G telecommunications and assessment design. Create a detailed scoring rubric for the following question based on the ground truth answer provided.

Question: {question}

Ground Truth Answer: {ground_truth}

Create a scoring rubric with a total of 5 points. Break down the points based on:
1. Core answer/concept (typically 2-3 points)
2. Key details/explanation (typically 1-2 points)  
3. Technical accuracy (typically 0.5-1 point)
4. Completeness (typically 0.5-1 point)

Return ONLY a JSON object with this structure:
{
  "total_points": 5,
  "criteria": [
    {"description": "criterion description", "points": X},
    {"description": "criterion description", "points": Y}
  ]
}

Make the rubric specific to this question. Do not include any other text outside the JSON.
\end{lstlisting}

After human curation and validation, we use the following prompt to score each answer with the question-specific rubrics:
\begin{lstlisting}
You are an expert grader for 5G telecommunications questions. Grade the predicted answer against the ground truth using the provided rubric.

Question: {question}

Ground Truth Answer: {ground_truth}

Predicted Answer: {predicted_answer}

Scoring Rubric (Total: {total_points} points):
{criteria_text}

Carefully evaluate the predicted answer against each criterion in the rubric. Award points based on:
- Factual correctness compared to ground truth
- Completeness of the answer
- Technical accuracy of terminology
- Whether key points from ground truth are covered

Return ONLY a JSON object with this exact structure:
{
  "score": X.X,
  "grading_reasoning": "Detailed explanation of the score, breaking down points awarded/deducted for each criterion"
}

The score should be a number between 0 and {total_points}, and can include decimals (e.g., 3.5, 4.0).
Be strict but fair. If the predicted answer contradicts the ground truth or misses key points, deduct points accordingly.
Do not include any text outside the JSON object.
\end{lstlisting}

\subsection{CR QA Generation Prompt}
\label{appendix:cr_qa_prompt}

We use the following prompt to generate question-answer pairs from approved Change Request documents. The prompt instructs the language model to create questions in one of two formats depending on the clarity of before-and-after states in the specification text, and to skip CRs that lack sufficient technical detail for meaningful QA generation.

\begin{lstlisting}
You are analyzing a 3GPP Change Request (CR) document. Based on the information provided, generate ONE question-answer pair.

CR Information:
- Specification: {spec_number}
- Summary of Change: {summary}
- Reason for Change: {reason}
- Consequences if Not Approved: {consequence}
- Clauses Affected: {clauses_affected}

Decision rule BEFORE producing output:
- If AND ONLY IF the Summary of Change, Reason for Change, AND Consequences are all very short one-liners (i.e., each is so brief that you cannot identify a meaningful technical change and rationale), then DO NOT create a Q&A. Instead, return:
  {
    "skip": true,
    "reason": "why the three fields are too simple to form a meaningful Q&A"
  }

Otherwise, create exactly ONE Q&A with the following instructions.

Instructions for Q&A:
1) Decide if you can identify clear "before" and "after" states in the spec text from the provided info.
2) Generate a question in ONE of these formats:

   Format A (if before AND after states are clear; do NOT mention the spec number in the question):
   - Example templates:
     * "Why was [specific parameter/behavior X] changed to [specific parameter/behavior Y]?"
     * "What is the reason for changing [feature/requirement X] to [feature/requirement Y]?"
     * "I observed that [X] was modified to [Y]. What is the rationale for this change?"

   Format B (if before OR after is unclear; explicitly mention the specification number):
   - Example templates:
     * "What is the reason for the change to [specific aspect] in specification {spec_number}?"
     * "Why was [feature/behavior] modified in 3GPP TS {spec_number}?"
     * "I observed a change regarding [topic] in specification {spec_number}. What is the reason?"

3) The answer should:
   - Explain the reason for the change,
   - Include the consequences if the change is not accepted,
   - Be natural and technical (no copy-paste; use proper terminology).

Output format (strict JSON):
- If skipping:
  {
    "skip": true,
    "reason": "short explanation"
  }

- If generating Q&A:
  {
    "question": "your generated question",
    "answer": "your generated answer",
    "format_used": "A" or "B"
  }
\end{lstlisting}

All generated question-answer pairs undergo manual review to verify technical accuracy, ensure natural phrasing, and confirm that answers adequately explain both the rationale for the specification change and the potential consequences if the change were not approved.

\subsection{Reference Helpfulness Filter}
\label{appendix:helpfulness_filtering}
We use the following prompt to filter the helpful references for the microbenchmark on cross-reference resolution:
\begin{lstlisting}
Compare the original chunk to the reference chunk.
original chunk: {org_chunk}
reference chunk: {ref_chunk}
Instructions:
1. See if the reference chunk provides information that would be useful to understand what the original chunk is talking about.
2. Return 'helpful' if the reference chunk's content will provide necessary detail to fully understand what the original chunk's content is talking about.
3. Return 'unhelpful' if the reference chunk's content is not necessary to fully understand what the original chunk is talking about.
4. Respond only with 'helpful' or 'unhelpful'  
\end{lstlisting}

\section{Additional Results on Real-World QA}
\label{appendix:real_world_qa}

\subsection{Dataset Categorization}
The 573 questions in our real-world QA dataset are organized into five technical categories that reflect the organizational structure of 5G standardization and engineering practice:
\begin{itemize}
    \item \textbf{Category 1: Air Interface \& Physical Layer} (186 questions, 32.5\%) -- Physical channel structures, modulation schemes, MIMO configurations, beamforming, and radio resource allocation in the NR air interface.
    \item \textbf{Category 2: Protocol Stack \& Resource Management} (174 questions, 30.4\%) -- Layer 2/3 protocol operations (MAC, RLC, PDCP, SDAP), quality-of-service mechanisms, numerology, and resource mapping procedures.
    \item \textbf{Category 3: Core Network \& Session Management} (114 questions, 19.9\%) -- 5G Core architecture, session establishment and mobility procedures, network function interactions, and signaling flows.
    \item \textbf{Category 4: Network Architecture \& Deployment} (66 questions, 11.5\%) -- Network slicing, multi-access architectures, deployment scenarios (NSA/SA), interworking with legacy systems, and infrastructure planning.
    \item \textbf{Category 5: Operations, Optimization \& Troubleshooting} (33 questions, 5.8\%) -- Performance optimization strategies, diagnostic procedures, measurement reporting, and operational best practices.
\end{itemize}

The distribution reflects the relative emphasis in 5G technical documentation and practitioner discussions, with lower-layer protocol mechanics and physical-layer operations receiving the most attention. Each question is assigned to exactly one category by the annotation team based on its primary technical focus.

\subsection{Additional Evaluation}
Table~\ref{tab:realworld_rubric_results} presents additional rubrics-scoring results. Table ~\ref{tab:category_results} presents the Per-category win rates against the GPT-4o baseline across five 5G technical domains, demonstrating the performance of base models, Chat3GPP, and \myname{} for different model families and sizes.

\begin{table}[t]
\centering
\small
\caption{Rubric-based scoring results comparing \myname{} against base models. Scores range from 0 to 5. All differences significant at p<0.0001 (Wilcoxon signed-rank test).}
\label{tab:realworld_rubric_results}
\begin{tabular}{lcc}
\toprule
\textbf{Metric} & \textbf{GPT-4.1} & \textbf{GPT-4.1-mini} \\
\midrule
Baseline Mean & 2.67 & 2.46 \\
\textbf{\myname{} Mean} & \textbf{3.29} & \textbf{3.06} \\
Improvement & +0.62*** & +0.60*** \\
Effect Size (dz) & 0.41 & 0.40 \\
\bottomrule
\end{tabular}
\end{table}

\section{Additional Results on CR-focused QA}
Table~\ref{tab:additional_cr} presents additional pairwise win rate results. Table~\ref{tab:evol_rubric} presents additional rubric-based scoring results.

\label{appendix:evol_full}
\begin{table}[t]
\centering
\small
\caption{Additional results on the CR-focused QA dataset.}
\resizebox{\columnwidth}{!}{
\begin{tabular}{llcc}
\toprule
\textbf{Model} & \textbf{Method} &
\makecell{\textbf{Length}\\\textbf{(tokens)}} &
\makecell{\textbf{Win Rate}\\\textbf{vs. GPT-4o (\%)}}
\\
\midrule
\multirow{3}{*}{Qwen3-4B}
  & Base Model   & 283 & 68.6 \\
  & Chat3GPP     & 372 & 81.4 \\
  & \textbf{\myname{}}    & 341 & \textbf{88.0} \\
\midrule
\multirow{3}{*}{Qwen3-8B}
  & Base Model   & 312 & 73.3 \\
  & Chat3GPP     & 466 & 85.7 \\
  & \textbf{\myname{}}    & 427 & \textbf{92.6} \\
\midrule
\multirow{3}{*}{Qwen3-14B}
  & Base Model   & 295 & 81.1 \\
  & Chat3GPP     & 481 & 91.4 \\
  & \textbf{\myname{}}    & 443 & \textbf{93.6} \\
\midrule
\multirow{3}{*}{Qwen3-32B}
  & Base Model   & 353 & 89.9 \\
  & Chat3GPP     & 549 & 93.7 \\
  & \textbf{\myname{}}    & 506 & \textbf{95.7} \\
\bottomrule
\end{tabular}
}
\label{tab:additional_cr}
\end{table}

\begin{table}[t]
\centering
\small
\caption{Rubric-based scoring results on CR-focused QA using GPT-4o as the base model. Scores range from 0 to 5. All pairwise differences significant at p<0.0001 (Wilcoxon signed-rank test).}
\label{tab:evol_rubric}
\begin{tabular}{lc}
\toprule
\textbf{Metric} & \textbf{Score} \\
\midrule
Base Model Mean & 2.26 \\
Chat3GPP Mean & 2.48 \\
\textbf{\myname{} Mean} & \textbf{3.05} \\
\midrule
\myname{} vs. Base & +0.79 (0.68) *** \\
\myname{} vs. Chat3GPP & +0.58 (0.51) *** \\
Chat3GPP vs. Base & +0.21 (0.26) *** \\
\bottomrule
\multicolumn{2}{l}{\small Effect sizes (dz) shown in parentheses.}
\end{tabular}
\end{table}

\section{Implementation Details}
\label{app:implementation}

\subsection{Inference Hyperparameters}
\label{app:hyperparams}

We adopt default or recommended settings for inference across APIs and locally deployed models:

\begin{itemize}
    \item \textbf{OpenAI models (GPT-4o, GPT-4.1):} temperature = 1.0, top\_p = 1.0
    \item \textbf{Claude 3.5 Haiku:} temperature = 1.0, top\_p = 1.0
    \item \textbf{Qwen3 models:} temperature = 0.6, top\_p = 0.95, top\_k = 20, with thinking mode enabled
\end{itemize}

\subsection{Computational Infrastructure}
\label{app:compute}

\paragraph{Hardware and API Access.}  
Experiments with GPT and Claude models were conducted via official APIs. For Qwen3 models, inference was performed locally using \texttt{vLLM} on a single NVIDIA A100 80GB GPU.

\subsection{Other Experimental Settings}
\label{app:defaults}

Unless otherwise specified:
\begin{itemize}
    \item \myname{} cross-reference resolution, the recursion depth is fixed at 2.
    \item All embeddings are obtained using \texttt{text-embedding-3-large}.
    \item GPT-4o serves as the default assistant, extractor, generator, and evaluator model.
\end{itemize}

\begin{table*}[t]
\centering
\small
\caption{Per-category performance on the real-world QA dataset (Win Rate vs. GPT-4o, \%).}
\resizebox{\textwidth}{!}{
\begin{tabular}{llccccc|c}
\toprule
\textbf{Model} & \textbf{Method} &
\makecell{\textbf{Air}\\\textbf{Interface}} &
\makecell{\textbf{Protocol}\\\textbf{Stack}} &
\makecell{\textbf{Core}\\\textbf{Network}} &
\makecell{\textbf{Network}\\\textbf{Arch.}} &
\makecell{\textbf{Operations \&}\\\textbf{Troubleshooting}} &
\textbf{Overall}
\\
\midrule
\multirow{3}{*}{GPT-4o}
  & Base Model   & (50) & (50) & (50) & (50) & (50) & (50) \\
  & Chat3GPP     & 54.8 & 67.8 & 64.5 & 62.1 & 69.7 & 62.4 \\
  & \textbf{\myname{}}    & \textbf{60.5} & \textbf{72.1} & \textbf{72.4} & \textbf{66.7} & \textbf{77.3} & \textbf{68.1} \\
\midrule
\multirow{3}{*}{Qwen3-4B}
  & Base Model   & 33.6 & 30.7 & 38.6 & 39.4 & 33.3 & 34.4 \\
  & Chat3GPP     & 59.9 & 70.4 & 71.9 & 72.0 & 66.7 & 67.3 \\
  & \textbf{\myname{}}    & \textbf{62.9} & \textbf{76.1} & \textbf{74.6} & \textbf{75.0} & \textbf{72.7} & \textbf{71.2} \\
\midrule
\multirow{3}{*}{Qwen3-8B}
  & Base Model   & 47.6 & 50.6 & 50.4 & 56.1 & 60.6 & 50.8 \\
  & Chat3GPP     & 72.0 & 78.4 & 71.9 & 74.2 & 75.8 & 74.4 \\
  & \textbf{\myname{}}    & \textbf{73.4} & \textbf{83.6} & \textbf{80.3} & \textbf{86.4} & \textbf{80.3} & \textbf{79.8} \\
\midrule
\multirow{3}{*}{Qwen3-14B}
  & Base Model   & 56.5 & 61.5 & 65.4 & 65.2 & 63.6 & 61.2 \\
  & Chat3GPP     & 73.9 & \textbf{85.9} & 79.8 & 75.8 & 77.3 & 79.1 \\
  & \textbf{\myname{}}    & \textbf{75.8} & 85.6 & \textbf{82.0} & \textbf{87.9} & \textbf{81.8} & \textbf{81.8} \\
\midrule
\multirow{3}{*}{Qwen3-32B}
  & Base Model   & 67.7 & 77.0 & 71.9 & 78.0 & 86.4 & 73.6 \\
  & Chat3GPP     & 79.6 & 87.4 & 84.2 & 78.0 & 83.3 & 82.9 \\
  & \textbf{\myname{}}    & \textbf{82.3} & \textbf{92.5} & \textbf{86.8} & \textbf{92.4} & \textbf{92.4} & \textbf{88.0} \\
\midrule
\multirow{3}{*}{\makecell[l]{Claude 3.5\\Haiku}}
  & Base Model   & 47.8 & 50.9 & 53.1 & 41.7 & 39.4 & 48.6 \\
  & Chat3GPP     & 51.6 & 62.4 & 57.5 & 53.0 & 59.1 & 56.6 \\
  & \textbf{\myname{}}    & \textbf{53.0} & \textbf{64.4} & \textbf{61.8} & \textbf{58.3} & \textbf{63.6} & \textbf{59.4} \\
\midrule
\multirow{3}{*}{GPT-4.1 mini}
  & Base Model   & 86.3 & 81.3 & 81.1 & 75.0 & 78.8 & 82.0 \\
  & Chat3GPP     & 84.4 & 89.1 & \textbf{89.9} & 88.6 & 90.9 & 87.8 \\
  & \textbf{\myname{}}    & \textbf{87.6} & \textbf{91.4} & 89.5 & \textbf{92.4} & \textbf{93.9} & \textbf{90.1} \\
\midrule
\multirow{3}{*}{GPT-4.1}
  & Base Model   & 89.2 & 87.6 & 83.8 & 84.8 & 89.4 & 87.2 \\
  & Chat3GPP     & 92.5 & 93.4 & 94.3 & \textbf{97.7} & 92.4 & 93.7 \\
  & \textbf{\myname{}}    & \textbf{93.8} & \textbf{95.4} & \textbf{96.1} & \textbf{97.7} & \textbf{97.0} & \textbf{95.4} \\
\bottomrule
\end{tabular}
}
\label{tab:category_results}
\end{table*}

\end{document}